\documentclass[letterpaper]{article} 
\usepackage{aaai24}  
\usepackage{times}  
\usepackage{helvet}  
\usepackage{courier}  
\usepackage[hyphens]{url}  
\usepackage{graphicx} 
\usepackage{natbib}  
\usepackage{caption} 
\usepackage{booktabs}       
\usepackage{amsfonts}       
\usepackage{nicefrac}       
\usepackage{microtype}      
\usepackage{xcolor}         
\usepackage{caption}
\usepackage{subcaption}
\usepackage{amsmath,amssymb,amsfonts}
\usepackage{algorithmic}
\usepackage{graphicx}
\usepackage{textcomp}
\usepackage{xcolor}
\usepackage{tabto}
\usepackage{comment}
\usepackage{diagbox}
\usepackage{flexisym}
\usepackage{bbding}
\usepackage{multirow}
\usepackage{makecell}
\usepackage{amsmath, bm}
\usepackage{tikz}
\usepackage{eso-pic}
\usepackage{algorithm}
\usepackage{threeparttable}

\newcommand{\bv}[0]{\ensuremath{\boldsymbol{b}} }

\newcommand{\hv}[0]{\ensuremath{\boldsymbol{h}} }
\newcommand{\iv}[0]{\ensuremath{\boldsymbol{i}} }

\newcommand{\xv}[0]{\ensuremath{\boldsymbol{x}} }
\newcommand{\yv}[0]{\ensuremath{\boldsymbol{y}} }
\newcommand{\zv}[0]{\ensuremath{\boldsymbol{z}} }

\newcommand{\Cv}[0]{\ensuremath{\boldsymbol{C}} }

\newcommand{\Hv}[0]{\ensuremath{\boldsymbol{H}} }
\newcommand{\Iv}[0]{\ensuremath{\boldsymbol{I}} }

\newcommand{\Kv}[0]{\ensuremath{\boldsymbol{K}} }

\newcommand{\Ov}[0]{\ensuremath{\boldsymbol{O}} }

\newcommand{\Qv}[0]{\ensuremath{\boldsymbol{Q}} }

\newcommand{\Tv}[0]{\ensuremath{\boldsymbol{T}} }

\newcommand{\Vv}[0]{\ensuremath{\boldsymbol{V}} }
\newcommand{\Wv}[0]{\ensuremath{\boldsymbol{W}} }
\newcommand{\Xv}[0]{\ensuremath{\boldsymbol{X}} }

\newcommand{\Zv}[0]{\ensuremath{\boldsymbol{Z}} }

\newcommand{\muv}[0]{\ensuremath{\boldsymbol{\mu}} }

\newcommand{\sigmav}[0]{\ensuremath{\boldsymbol{\sigma}} }

\usepackage{newfloat}
\usepackage{listings}
\DeclareCaptionStyle{ruled}{labelfont=normalfont,labelsep=colon,strut=off} 
\floatstyle{ruled}
\newfloat{listing}{tb}{lst}{}
\floatname{listing}{Listing}
\begin{document}
%
\title{Considering Nonstationary within Multivariate Time Series with Variational Hierarchical Transformer for  Forecasting}
\author{
    Muyao Wang,
    Wenchao Chen \thanks{Correspondence to Wenchao Chen (chenwenchao@xidian.edu.cn)},
    Bo Chen
}
\affiliations{
   National Laboratory of Radar Signal Processing Xidian University, Xi’an, Shaanxi, China \\
   muyaowang@stu.xidian.edu.cn, chenwenchao@xidian.edu.cn, bchen@mail.xidian.edu.cn

}
\maketitle
\begin{abstract}


The forecasting of {\bf M}ultivariate {\bf T}ime {\bf S}eries (MTS) has long been an important but challenging task.
Due to the non-stationary problem across long-distance time steps, previous studies primarily adopt stationarization method to attenuate the non-stationary problem of the original series for better predictability. However, existing methods always adopt the stationarized series, which ignores the inherent non-stationarity, and has difficulty in modeling MTS with complex distributions due to the lack of stochasticity. To tackle these problems, we first develop a powerful hierarchical probabilistic generative module to consider the non-stationarity and stochastity characteristics within MTS, and then combine it with transformer for a well-defined variational generative dynamic model named {\bf H}ierarchical {\bf T}ime series {\bf V}ariational {\bf Trans}former (HTV-Trans), which recovers the intrinsic non-stationary information into temporal dependencies. 
Being a powerful probabilistic model, HTV-Trans is utilized to learn expressive representations of MTS and applied to forecasting tasks. Extensive experiments on diverse datasets show the efficiency of HTV-Trans on MTS forecasting tasks. 

\end{abstract}

\section{Introduction}
\begin{figure}
\includegraphics[height =3.5cm,width = 8cm]{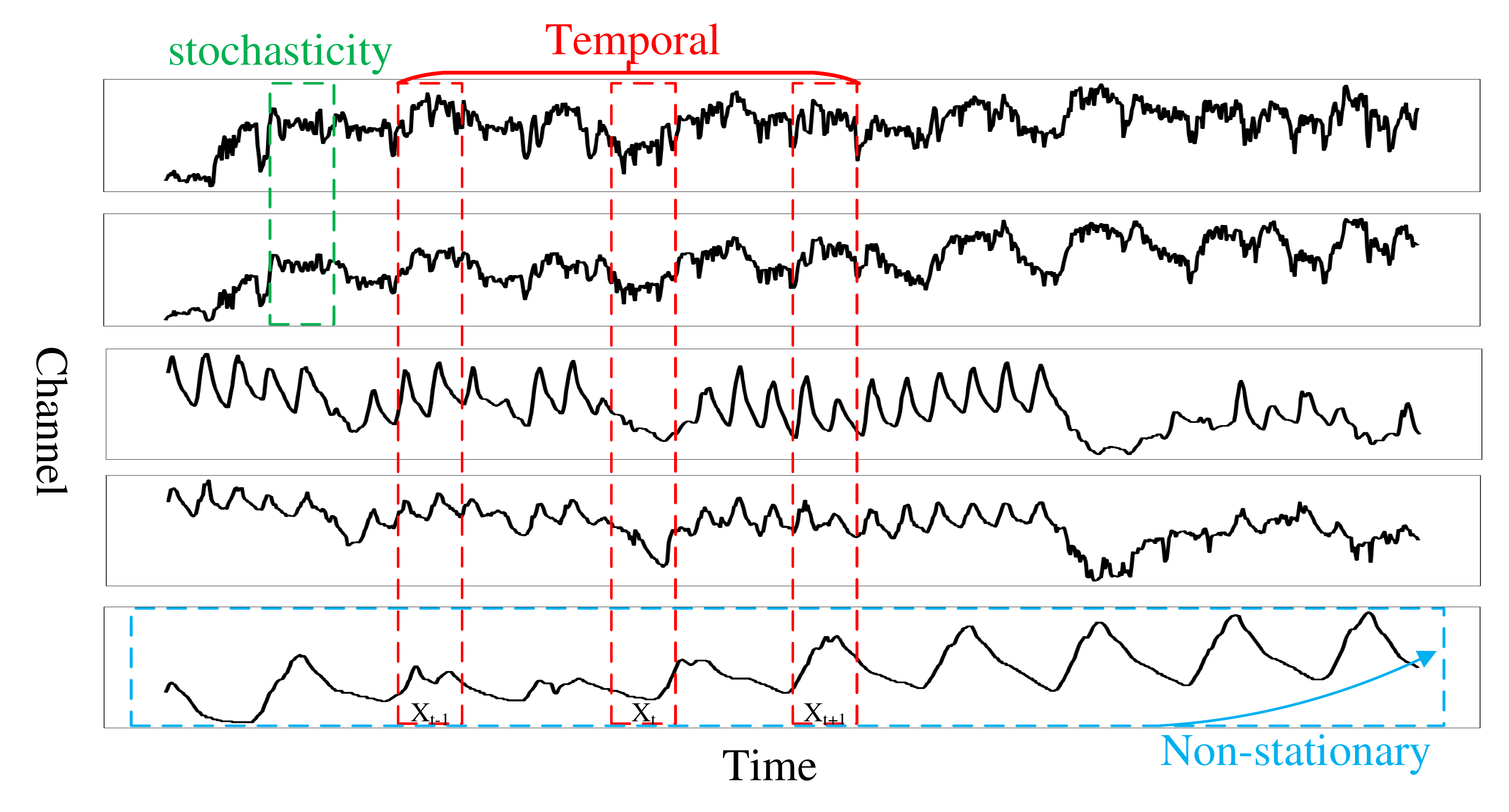}
\caption{{The temporal dependency, stochasticity and non-stationarity within MTS.}}\label{fig:motivation}
\end{figure}

Multivariate time series (MTS) is an important type of data that 
arises from a wide variety of domains, including internet services
, industrial devices
, health care
and finance
, to name a few. However,
the forecasting of MTS has always been a challenging problem as there exists 
not only complex temporal dependencies, as shown in the red box in Fig.~\ref{fig:motivation}, but also inherently stochastic components, as shown in the green box in Fig.~\ref{fig:motivation}.  Moreover, there exist non-stationary issues, as shown in the blue box in Fig.~\ref{fig:motivation}, which haves a huge impact on predictive performance.
To model the temporal dependencies of MTS, many dynamic methods based on recurrent neural networks (RNNs) have been developed  \cite{malhotra2016lstm,zhang2019deep,bai2019passenger,tang2020joint,yao2018deep}. 
Meanwhile, to consider the stochasticity, some probabilistic dynamic methods have also been developed  \cite{dai2021sdfvae,dai2022switching,chen2020switching,chen2022infinite,salinas2020deepar}.
With the development of Transformer \cite{vaswani2017attention} and due to its ability to capture long-range dependencies 
\cite{wen2022transformers,dosovitskiy2021image,chen2021autoformer}, 
and interactions,
which is especially attractive for time series forecasting, there is a recent trend to construct Transformer based MTS forecasting methods and have achieved promising results in learning expressive representations for MTS forecasting tasks. 
Recently, there are lots of efficient Transformer-based forecasting methods, such as Autoformer \cite{wu2021autoformer}, FEDformer \cite{zhou2022fedformer}, Pyraformer \cite{liu2021pyraformer}, Crossformer \cite{zhang2023crossformer} and so on. 
Despite the advanced architectural design, it is still hard for Transformers to predict real-world time series because of the non-stationarity of data.
Non-stationary time series is characterized by the continuous change of statistical properties and joint distribution over time, making them less predictable \cite{hyndman2018forecasting}. In previous works, they usually pre-process the time series by stationarization \cite{ogasawara2010adaptive,passalis2019deep,kim2021reversible}, which can attenuate the non-stationarity of raw time series and provide more stable data distribution for deep models. However, these MTS stationarization methods ignore the non-stationarity of real-world series, which will result in the over-stationarization problem \cite{liu2022nonstationary}. 

Non-stationary transformer \cite{liu2022nonstationary} is proposed to tackle the over-stationarization problem in the Transformers by approximating distinguishable attentions learned from raw series, which is a method limited to a specific transformer situation. 
Moreover, because of the deterministic architectural designs, the methods mentioned before still face challenges when it comes to predicting real-world time series due to the non-deterministic data caused by noise data. 
Therefore, in this paper, we analyze the series forecasting task from another point of view to bring the non-deterministic and non-stationarity back to the deep models, especially for the transformer, and propose a novel dynamic hierarchical generative module to effectively capture both inherent properties.

Moving beyond the constraints of previous work, considering the non-determinism and non-stationarity within MTS and enhancing the representative power of deep models, we develop a {\bf H}ierarchical {\bf T}ime series {\bf V}ariational  {\bf Trans}former (HTV-Trans), which is a well-defined probabilistic dynamic model obtained by combining a proposed {\bf H}ierarchical {\bf T}ime series {\bf P}robabilistic {\bf G}enerative {\bf M}odule (HTPGM), as illustrated in Fig.~\ref{fig:generate} (a)(b), with the Transformer block, as in  Fig.~\ref{fig:generate} (c). Specifically, HTPGM module is able to get the multi-scale non-deterministic and non-stationary representations of the original MTS, which can be served to guide the forecasting tasks with transformer block.
In addition, we introduce an autoencoding variational inference scheme for efficient inference and a joint optimization objective that combines forecasting and reconstruction loss to recover the non-deterministic and non-stationary time-series representation into Transformer. The main contributions of our work are summarized as follows:
\begin{itemize} 
    \item For MTS forecasting, we propose HTPGM module, which is able to consider the non-stationary information within the MTS with a hierarchical multi-scale generative structure, thus avoiding the over-stationarization problem.
    \item We develop HTV-Trans, a probabilistic dynamic model with  HTPGM as a generative module, which can consider the non-deterministic and non-stationary issues within the temporal dependencies of MTS. 
    \item For optimization, we introduce an autoencoding inference scheme with a combined prediction and reconstruction loss to enhance the representation power of MTS.
    \item Experiments on different datasets illustrate the efficiency of our model on MTS forecasting task.
\end{itemize}

\section{Backgrounds}
\subsection{Multivariate Time Series Forecasting}
In recent years, transformer models have subsequently emerged and have shown great power in sequence modeling. To overcome the quadratic computational growth in relation to sequence length, subsequent works have aimed to reduce the complexity of Self-Attention. In particular, for time series forecasting, Informer\cite{zhou2021informer} extends Self-Attention with a KL-divergence criterion to select dominant queries. Reformer\cite{kitaev2019reformer} introduces local-sensitive hashing (LSH) to approximate attention by allocating similar queries. Not only have these models been improved by reduced complexity, they have also developed more complex building blocks for time series forecasting. Autoformer\cite{wu2021autoformer} fuses decomposition blocks into a canonical structure and develops Auto-Correlation to discover series-wise connections. Pyraformer \cite{liu2021pyraformer} designs a pyramid attention module (PAM) to capture temporal dependencies at different hierarchies. Other deep Transformer models have also achieved remarkable performance. Fedformer\cite{zhou2022fedformer} designs two attention modules, which use Fourier transform and wavelet transform to process attention operations in the frequency domain. Non-stationary transformer\cite{liu2022nonstationary} designs a de-stationary attention, which approximates the distinguishable attention in non-stationary sequences to solve the problem of excessive stationarity. Crossformer\cite{zhang2023crossformer} proposes a Two-Stage-Attention (TSA) layer to capture the cross-time and crossdimension dependency of the embedded array and show its effectiveness over previous state-of-the-arts. In this paper, we take a different approach from previous works that focus on transformer architectural design. In this paper, we propose a novel approach to address the non-stationary and non-deterministic properties of time series, which are essential characteristics of this type of data \cite{kim2021reversible,liu2022nonstationary}. 

\subsection{Non-stationary Problems of Time Series }
While stationarity is important to the predictability of time series \cite{kim2021reversible,liu2022nonstationary}, real-world series always
present non-stationarity. To tackle this problem, the classical statistical method ARIMA \cite{gepbox1976Time}
stationarizes the time series through differencing. As for deep models, since the distribution-varying
problem accompanied by non-stationarity makes deep forecasting even more intractable, stationarization methods are widely explored and always adopted as the pre-processing for deep model inputs.
Adaptive Norm \cite{eduardo2010Adaptive} applies z-score normalization for each series fragment by global statistics of a
sampled set. DAIN \cite{npassa2019Deep} employs a nonlinear neural network to adaptively stationarize time series
with observed training distribution. RevIN \cite{kim2021reversible} introduces a two-stage instance normalization that transforms model input and output respectively to reduce the discrepancy of each series, which brings great benefit to the model’s capability of modeling non-stationary time series. 
However, Non-stationary transformer \cite{liu2022nonstationary} find out that directly stationarizing time series will damage the model’s capability of modeling specific temporal dependency, which is named over-stationarization and tackle the problem by approximating distinguishable attentions learned from the original series.
Given this over-stationarization issue, unlike Non-stationary-transformer method, HTV-Trans further develops a hierarchical generative module to capture the multi-scale statistical characteristics of the original input time series, which can bring more intrinsic non-stationarity of the original series back to latent representation for time series forecasting.

\begin{figure}[!t]
\centerline{\includegraphics[width=1\linewidth,height=0.55\linewidth]{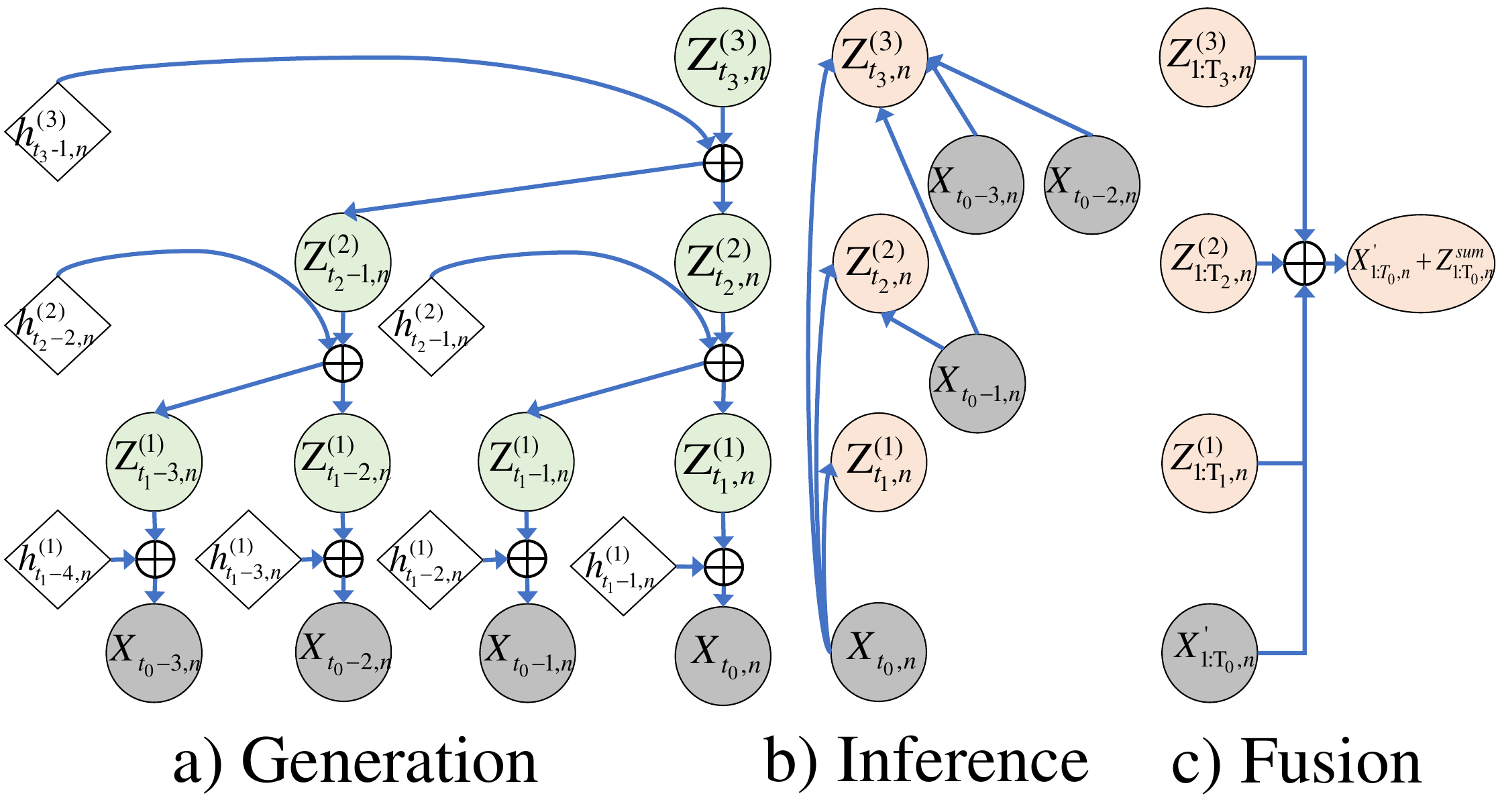}}
\caption{Graphical illustration of different operations of the HTV-Trans: (a) generative process of HTPGM, (b) the inference scheme of HTPGM. (c) the fusion of different scale information and stationarization input series for forecasting.} 
\label{fig:generate}
\end{figure}
\section{Method}
\subsection{Problem Definition}
Defining the $n$-th MTS as $\xv_n=\{\xv_{1,n}, \xv_{2,n}, ..., \xv_{T,n}\}$ , where $n=1,...N$ and $N$ is the number of MTS. $T$ is the duration of $\xv_n$ and the observation at time $t$, $\xv_{t,n} \in \mathbb{R}^{V}$, is a $V$ dimensional vector where $V$ denotes the number of channels, thus $\xv_n \in \mathbb{R}^{T \times V}$. 
\subsection{Time Series Stationarization} \label{embedding}
Non-stationary time series present a significant challenge for deep models in forecasting tasks, as they tend to struggle with generalization to series exhibiting variations in mean and standard deviation during forecasting. This is due to the inherent difficulty of modeling time series with changing statistical properties. The pilot work, RevIN \cite{kim2021reversible} and Non-stationary transformer \cite{liu2022nonstationary} both put forward the instance normalization to each input and restores the statistics to the corresponding output in a similar way, which makes each series follow a similar distribution. Empirical evidence has shown that this design is effective, but Non-stationary transformer introduces an alternative method called Series Stationarization that requires fewer computational resources. Based on these considerations, we adopt Series Stationarization for normalizing our input data.
\\\
{\bf Normalize module:} To attenuate the non-stationarity of each input series, Series Stationarization conducts normalization on the temporal dimension by a sliding window over time.
Here are the equations.
\begin{equation} \textstyle \label{normalization}
\begin{array}{l}
\begin{split}
&\muv_{\mathbf{\xv}}=\frac{1}{T} \sum_{i=1}^{T} \xv_{i}, \sigmav_{\mathbf{x}}^{2}=\frac{1}{T} \sum_{i=1}^{T}\left(\xv_{i}-\muv_{\mathbf{x}}\right)^{2} \\
&\xv_{i}^{\prime}=\frac{1}{\sigmav_{\mathbf{x}}} \odot\left(\xv_{i}-\muv_{\mathbf{x}}\right) \\
\end{split}
\end{array}
\end{equation}
where $\muv_{\mathbf{\xv}}, \sigmav_{\mathbf{\xv}} \in \mathbb{R}^{V \times 1}$,$\frac{1}{\sigmav_{\mathbf{x}}}$
means the element-wise division and $\odot$ is the element-wise product. It is obvious that the Series Stationarization aims to reduce the distributional differences among individual input time series, thereby stabilizing the distribution of model inputs.\\\
{\bf Denormalize module:}
After the base model $\Hv$ predicting the future value with length-$\Tv$, we adopt De-normalization to transform the model output $\yv^{\prime} = [\yv_{1}^{\prime},\yv_{2}^{\prime},...,\yv_{t}^{\prime}]\in \mathbb{R}^{T \times V}$ with $\sigmav_{x}$ and $\mu_{x}$ , thus obtain $\yv = [\yv_{1},\yv_{2},...,\yv_{t}]$ as the eventual forecasting results. The De-normalization module can be formulated as follows: \begin{equation} \textstyle \label{normalization}
\yv^{\prime}=\mbox{\Hv}(\xv^{\prime}), \yv_{i}=\sigmav_{\xv} \odot\left(\yv_{i}^{\prime}+\muv_{\mathbf{\xv}}\right)
\end{equation}
Because of the two-stage transformation, the base models will receive stationarized inputs, which follow a stable distribution and are easier to generalize. These designs also make the model equivariant to translational and scaling perturbance of time series, which benefits real-world series forecasting.

\subsection{Hierarchical Time Series Variational Transformer}
{\bf{Analysis of over-stationarization: }}Although the statistics of each output time series are explicitly restored to the corresponding original distribution of input series by De-normalization, the non-stationarity of the original series cannot be fully restored only in this way. According to the conclusion from the Non-stationary transformer\cite{liu2022nonstationary}, the undermined effects caused by over-stationarization happen inside the deep model, especially in the calculation of attention. Furthermore, non-stationary time series are fragmented and normalized into several series chunks with the same mean and variance, which follow more similar distributions than the raw data before stationarization. Thus, the model is more likely to generate over-stationary and uneventful outputs, which is irreconcilable with the natural non-stationarity of the original series.
To address the over-stationarization issue caused by Series Stationarization, we propose a novel Hierarchical Time series Variational Autoencoder(HTPGM) that can solve this problem more completely.
Similar to the Non-stationary transformer, we fuse the normalized data and the non-stationary information (as shown in  Fig.~\ref{fig:generate} (c))  provided by the hierarchical distribution to solve the problem of similar attention caused by the similar distribution of the input data. Furthermore, our hierarchical generative module can provide fine-grained data distribution information through the different hidden layers distribution at different scales, which can boost the non-stationary series predictive capability of the base model.  
\begin{figure}[!t]
\centerline{\includegraphics[width=1\linewidth,height=0.45\linewidth]{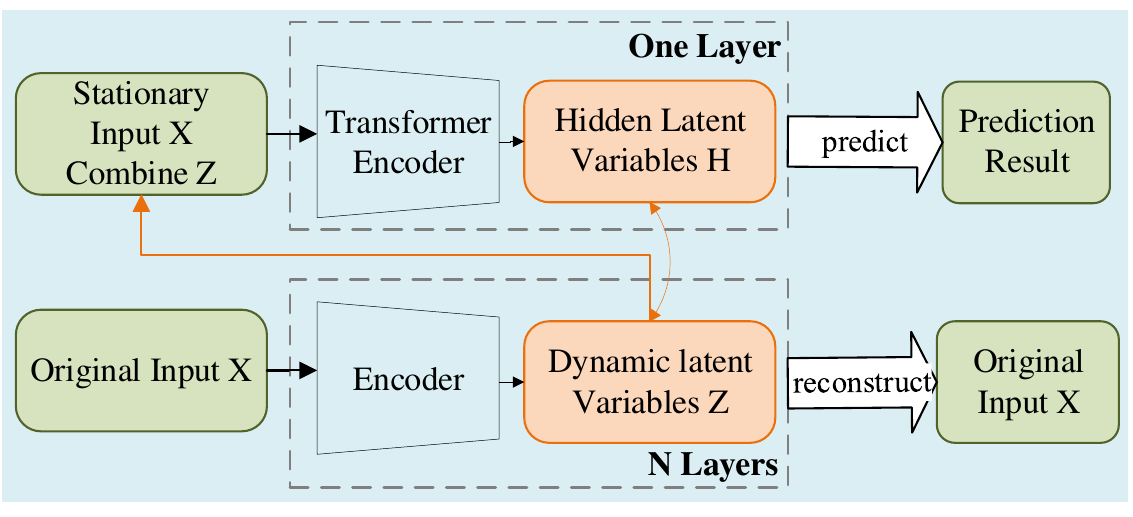}}
\caption{the whole framework of HTV-Trans.} 
\label{fig:framework_HV-trans}
\end{figure}
\\\
{\bf{Hierarchical generative module: }}As previously mentioned, to tackle the over-stationarization problem caused by Series Stationarization, Non-stationary transformer proposes a novel De-stationary Attention mechanism, which focuses on the attention caused by stationarization and discovers the particular temporal dependencies from original non-stationary data. From another point of view, we introduce a hierarchical probabilistic generative module to consider the original series statistical information through the generative process in our proposed model. The whole generative module is illustrated in Fig.~\ref{fig:generate} (a). This module aims to consider the inherent uncertainty and intrinsic non-stationarity of the original input time series and enhance the capacity of the model to handle complex, non-deterministic, and non-stationary time series distributions. 
Our model mainly starts with three aspects to address the issues which are mentioned previously.\\\
{\bf Vanished non-stationary information supplement: }Firstly, given the intrinsic non-stationarity of input series, we consider that the post distribution of HTPGM should depend on the original input series as shown in Fig.~\ref{fig:generate}. Due to the reconstruction of raw input series, the intrinsic non-stationarity can be learnable through the generative process. \\\
{\bf Multi-scale non-stationary information: }Secondly, we design a generative process in which hidden layer distribution has different scales for HTPGM. The time dimension of hidden variables $\zv_{i}$ from the bottom layer to the top layer is gradually reduced. Our design aims to let the distribution of different scales capture more fine-grained original time series information. Because the way of information fusion is achieved by using the nearest neighbor interpolation to unify the dimension of hidden layer distribution, this design is equivalent to setting several windows to capture the distribution in different small chunks of raw time series. 
Due to the different scales of hidden layer distribution, the hierarchical generative process can provide fine-grained distribution information, which is stored in $\zv_{sum}$ as shown in Fig.~\ref{fig:generate} (c). \\\
{\bf Robust representation: }Our proposed generative module is also capable of solving the non-deterministic problems of time series by utilizing the probabilistic component. The probabilistic component enables the incorporation of non-determinism into the representation of time series, by introducing a degree of randomness and uncertainty in the generative process. This characteristic of the probabilistic generative module makes it robust to the noise in the input, allowing for the modeling of non-deterministic time series. 
Overall, our generative module can extract a non-stationary and roust representation of time series.\\\
 {\bf{Generation:}} 
 We propose a special way that incorporates the output of transformer $\hv_{\mathbf{t}}$(as shown in Fig.~\ref{fig:generate} (a)) into the generative process of HTPGM. The generative process is illustrated in Fig.~\ref{fig:generate} (a). This operation enhances the ability of $\hv_{\mathbf{t}}$ and latent variable $\zv_{\iv}$ to reconstruct the original input series, resulting in an expressive and underlying non-stationary representation of the time series. The detail of the generative process is defined as
\begin{equation}
\begin{split} \label{generation}
& \zv_{{t_L},n}^L \sim {\cal N}\left( {\Wv_{{t_{L}},n}^{L}\hv_{{t_{L}},n}^{L},\Iv} \right)\ 
, \quad \cdots \cdots \\ 
& \zv_{{t_i},n}^i,...,\zv_{{t_i} + {K_i},n}^i \sim {\cal N}\left( {\muv _{{t_{i + 1}},n}^{i + 1}\left( {\zv_{{t_{i + 1}},n}^{i + 1}} \right),{\mathop{\mbox {diag}}\nolimits} \left( {\sigmav _{}^i} \right)} \right) \\ 
& \muv _{{t_{i + 1}},n}^{i + 1}\left( {\zv{{t_{i + 1}},n}^{i + 1}} \right) = f(\Wv_{\zv,\muv }^{i + 1}\zv_{t - 1,n}^{i + 1} + \Wv_{h\mu }^i\hv_{{t_{i + 1}} - 1,n}^{i + 1}) \\ 
& \cdots \cdots, \quad \xv_{{t_0},n}^{} \sim {\cal N}\left( {\muv_{t,n}^1,{\mathop{\mbox {diag}}\nolimits} \left( {\sigmav _{}^x} \right)} \right), \\ 
& \muv _{t,n}^1 = f\left( {\Wv_{z,\mu }^{i + 1}z_{t - 1,n}^{i + 1} + \Wv_{h\mu }^1\hv_{{t_{i + 1}} - 1,n}^{i + 1}} \right)  \nonumber
\end{split}
\end{equation}
where $ \Wv_{\zv\mu}^i, \Wv_{\hv\mu}^i \in \mathbb {R}^{K^i \times V}$, are all learnable parameters of the generate network, $K^i$, $t_i$ are changed with different layer i. $f(\cdot)$ refers to the non-linear activation function. $\muv _{t,n}^i$ and $\sigmav _{t,n}^i$ are means and covariance parameters of $\zv_{t,n}^i$. $\hv_{t-1,n} \in {\mathbb R}^{T \times V}$ denotes the deterministic latent states of our forecasting module. We combine $\zv_{t,n}^i$ and $\hv_{t-1,n}$ into the generative process to consider the temporal dependencies and the stochasticity. The prior information is passed down through the generative process from the top layer to the bottom layer of the HTPGM. This hierarchical generative process allows each layer of HTPGM to learn the non-stationary information from different perspectives. It enables the model to capture a diverse range of temporal dependencies and variations within the original input time series, resulting in a more robust and non-stationary multi-scale representation of the original series. This hierarchical approach allows the model to effectively handle complex, non-stationary time series, leading to improved performance in forecasting tasks.
\\\
{\bf{Hierarchical variational transformer:}}
In order to maximize the advantages of our hierarchical generative module, we have designed a transformer-based prediction model, which we call HTV-trans (as shown in Fig.~\ref{fig:framework_HV-trans}). In this prediction model, we use a one-layer transformer encoder as our model basic feature extractor and a MLP as a forecasting block to output the prediction in one step.
\\\
{\bf{Transformer block:}} 
Given the long time series dependency, we choose a transformer encoder to capture the dynamic information from the normalized input series. As mentioned previously, we consider that the $\zv_{sum}$ as shown in Fig.~\ref{fig:generate} (c) can store the vanished non-stationary information of original time series. Thus, we decide to integrate the normalized series and $\zv_{sum}$ as input to solve the over stationarizaiton problem.
Specifically, we introduce a {\bf M}ulti-head {\bf S}elf {\bf A}ttention (MSA) block to capture the temporal dependence as
\begin{equation}\begin{split} \label{infer-1} \textstyle
&\Ov = \mbox{MSA}(\mbox{Embedding}(\Xv ^{'})+ \\
&\qquad \alpha\mbox{sum}(\mbox{Interpolate}(\Zv_i))) = {\mathop{\mbox {con}}\nolimits} \left( {{{\Hv}_1}, \ldots,{{\Hv}_m}} \right)\\
&{{\Hv}_i} = \mbox {SA}\left( {{\Qv_i},{\Kv_i},{\Vv_i}} \right){ = }{\mathop{\mbox {softmax}}\nolimits} ( \textstyle{\frac{{{\Qv_i^T}{\Kv_i}}}{{\sqrt {{d_K}} }}} ){\Vv_i}\\
\end{split}
\end{equation}
where $\Xv^{'} \in \mathbb {R}^{V \times T}$ , denotes the normalized input, $\mbox{Embedding}(\cdot)$ denotes the time feature embedding which is mentioned in \cite{zhou2021informer}, sum($Z$) denotes the sum of the hidden layers of HTPGM and $\mbox {con}(\cdot)$ means concatenate operation,
${\Qv_i} = \Wv_Q^i(\Xv^{'}+\mbox{sum}(\mbox{Interpolate}(\Zv))) \in \mathbb{R}^{d_k \times T},{\Kv_i} = \Wv_K^i(\Xv^{'}+{\mbox{sum}}(\mbox{Interpolate}(\Zv))) \in \mathbb{R}^{d_k \times T}$, where $i \in \{ 1,2,...,m \}$ and $m$ is the number of heads. $\Ov \in \mathbb{R}^{(V \times T \times  m)}$ is the output of MSA block. It is important to note that we introduce a new parameter $\alpha$, which decides the balance between the stationary information and non-stationary information. To prove the validity of $\alpha$, we introduce an ablation study of $\alpha$ in section 5. 
After combining temporal, non-stationary and non-deterministic information of MTS into $\tilde \hv \in \mathbb{R}^{V \times T}$, feed forward network blocks are further applied for exploring expressive representations of MTS and getting the dynamic latent states $\hv \in \mathbb{R}^{V \times T}$.\\\
{\bf{MLP for forecasting: }}Different from other transformer models, we select the MLP decoder for forecasting tasks due to the fact that the traditional transformer decoder tends to fit time series data too rapidly, leading to inadequate learning of effective representations in the latent layers of HTPGM. After a number of experiments, we find that the MLP decoder is more appropriate for our generative module. In order to output the prediction in one step, the MLP output size is set to be the same as the prediction length. 
Combining HTPGM and Transformer, we finally develop HTV-Trans, a novel hierarchical probabilistic generative dynamic model. The graphical illustration of the whole framework is shown in Fig.~\ref{fig:framework_HV-trans}.\\\
{\bf{Multi-scale inference network: }}Following VAE-based models, we define a Gaussian distributed variational distribution $q({\zv_{t,n}}) = {\cal N}(\muv _{t,n}^{},\mbox{diag}(\sigmav _{t,n}))$ to approximate the true posterior distribution $p(\zv_{t,n}|-)$, and map the  dynamic input series ${{\hat \xv}_{t,n}}$ to their parameters as:
\begin{equation}
\begin{split}\label{inference}
&q({\zv_{t,n}}) = {\cal N}(\muv _{t,n}^{},\mbox{diag}(\sigmav _{t,n}))\\
&\muv _{t,n}^{i}(\xv) = f\left( {\Cv_{x\mu }^{i}{\left({\xv} \right)} + {\bv_{x\mu }^{i}}} \right)\\
&\sigmav_{t,n}^{i}(\xv) = {\mbox {Softplus}}\left( {f\left( {\Cv_{x\sigma }^{i}{\left({\xv}\right)} + {\bv_{x\sigma }^{i}}} \right)} \right) \\
 \end{split}
\end{equation}
where $\Cv_{x\mu }^{i}, \Cv_{x\sigma }^{i} \in \mathbb {R}^{K \times V^{'}}$, $\bv_{x\mu }^{i}, \bv_{x\sigma} ^{i} \in \mathbb {R}^{K}$ are all learnable parameters of the inference network.$f(\cdot)$ refers to the non-linear activation function.  
Based on the structure of the inference network as shown in Fig.~\ref{generation} (b), the posterior of probabilistic latent variables of HTPGM are approximated by multi-scale original input series, thus enabling richer latent representations for HTPGM.

\begin{table*}[!t]
\begin{center}
    \small
    \setlength{\tabcolsep}{1.8mm}
\begin{tabular}{cccccccccccccccc}
\toprule
\multicolumn{2}{c}{\multirow{2}{*}{\begin{tabular}[c]{@{}c@{}}Models\\ Metric\end{tabular}}} & \multicolumn{2}{c}{HTV-Trans} & \multicolumn{2}{c}{Autoformer} & \multicolumn{2}{c}{Informer} & \multicolumn{2}{c}{NS Transformer} & \multicolumn{2}{c}{Fedformer} & \multicolumn{2}{c}{Pyrafomer} & \multicolumn{2}{c}{Crossformer} \\ \cmidrule{3-16} 
\multicolumn{2}{c}{}                                                                         & MSE           & MAE          & MSE            & MAE           & MSE           & MAE          & MSE           & MAE          & MSE           & MAE          & MSE          & MAE     & MSE          & MAE    \\ \midrule
\multicolumn{1}{c|}{\multirow{4}{*}{ETTh1}}             & \multicolumn{1}{c|}{96}            & {0.389}              & \bf{0.396}             & 0.536           & 0.548         & 0.984     & 0.786        & 0.513              &  0.491            &  \bf{0.376}       & 0.419        & 0.783             &  0.657    &0.431 &  0.441     \\
\multicolumn{1}{c|}{}                                   & \multicolumn{1}{c|}{192}           & {0.445}        & \bf{0.422}        & 0.543          & 0.551         & 1.027         & 0.791        & 0.534         & 0.504        & {0.420}         & 0.448        & 0.863        & 0.709   & \bf{0.411}  & 0.440   \\
\multicolumn{1}{c|}{}                                   & \multicolumn{1}{c|}{336}           & {0.487}         & \bf{0.440}        & 0.615          & 0.592         & 1.032         & 0.774        & 0.588         & 0.535        & {0.459}         & 0.465        & 0.941        & 0.753  & \bf{0.441} & 0.461     \\
\multicolumn{1}{c|}{}                                   & \multicolumn{1}{c|}{720}           & \bf{0.489}         & \bf{0.455}        & 0.599          & 0.600         & 1.169      & 0.858        &  0.643             &  0.616            & 0.506        & 0.507        &   1.042          &     0.819 &0.515 &0.518       \\ \midrule
\multicolumn{1}{c|}{\multirow{4}{*}{ETTh2}}             & \multicolumn{1}{c|}{96}            & \bf{0.300}              & \bf{0.338}              & 0.492           & 0.517         &  2.826       & 1.330        &   0.476            & 0.458              & 0.346        & 0.388        &  1.380            &    0.943  &0.860 &0.691       \\
\multicolumn{1}{c|}{}                                   & \multicolumn{1}{c|}{192}           & \bf{0.382}         & \bf{0.391}        & 0.556          & 0.551         & 6.186         & 2.070        & 0.512         & 0.493        & 0.429         & 0.439        & 3.809        & 1.634  &1.026 &0.729    \\
\multicolumn{1}{c|}{}                                   & \multicolumn{1}{c|}{336}           & \bf{0.377}         & \bf{0.405}        & 0.572          & 0.578         & 5.268         & 1.942        & 0.552         & 0.551        & 0.482         & 0.480        & 4.282        & 1.792 &1.110 &0.768      \\
\multicolumn{1}{c|}{}                                   & \multicolumn{1}{c|}{720}           & \bf{0.412}         & \bf{0.438}        & 0.580          & 0.588         & 3.667      & 1.616        &  0.562             &   0.560           & 0.463        & 0.474        &   4.252           &    1.790  &2.151 & 1.134      \\ \midrule
\multicolumn{1}{c|}{\multirow{4}{*}{ETTm1}}             & \multicolumn{1}{c|}{96}       & {0.337}         & \bf{0.354}        & 0.523          & 0.488         & 0.615         & 0.556        & 0.386         & 0.398        & 0.378         & 0.418        & 0.536        & 0.506   &\bf{0.320} &0.373    \\
\multicolumn{1}{c|}{}                                   & \multicolumn{1}{c|}{192}           & \bf{0.366}         & \bf{0.374}        & 0.543          & 0.498      & 0.723          & 0.620       & 0.459              &  0.444            & 0.426         & 0.441       &  0.539            &      0.520  &0.407 &0.437     \\
\multicolumn{1}{c|}{}                                   & \multicolumn{1}{c|}{336}           & \bf{0.412}              & \bf{0.396}             &  0.675    &  0.551        & 1.300          & 0.908        &   0.495            & 0.464            & 0.445        & 0.459        &    0.720          &   0.635  &0.417 &0.433     \\
\multicolumn{1}{c|}{}                                   & \multicolumn{1}{c|}{720}           & \bf{0.484}         & \bf{0.434}        & 0.720          & 0.573         & 0.972         & 0.744
& 0.585         & 0.516        & 0.543         & 0.490        & 0.940        & 0.740 &0.610 &0.554       \\ \midrule
\multicolumn{1}{c|}{\multirow{4}{*}{ETTm2}}             & \multicolumn{1}{c|}{96}            &\bf{0.178}         & \bf{0.259}        & 0.255          & 0.339         & 0.365         & 0.453        & 0.192         & 0.274        & 0.203         & 0.287        & 0.409        & 0.488  &0.490 &0.487     \\
\multicolumn{1}{c|}{}                                   & \multicolumn{1}{c|}{192}           & \bf{0.248}         & \bf{0.301}        & 0.281          & 0.340         & 0.533      & 0.563        &   0.280            &   0.339           & 0.269        & 0.328        &   0.673          &  0.641  & 0.922 &0.711   \\
\multicolumn{1}{c|}{}                                   & \multicolumn{1}{c|}{336}           & \bf{0.311}              & \bf{0.341}             &  0.339    & 0.372          & 1.363          & 0.887        &  0.334             &  0.361            & 0.325        & 0.366         &   1.210          &  0.846   & 0.770 & 0.590        \\ 
\multicolumn{1}{c|}{}                                   & \multicolumn{1}{c|}{720}           & \bf{0.405}         & \bf{0.394}        & 0.422          & 0.419         & 3.379         & 1.388        & 0.417        & 0.413        & 0.421         & 0.415        & 4.044        & 1.526 &0.920 &0.730    \\ \midrule
\multicolumn{1}{c|}{\multirow{4}{*}{Weather}}                                   & \multicolumn{1}{c|}{96}            & {0.181}         & \bf{0.223}        & 0.266          & 0.336         & 0.300         & 0.384        & 0.173         & 0.223        & 0.217         & 0.296        & 0.354        & 0.392   & \bf{0.163} &0.226    \\
\multicolumn{1}{c|}{}           & \multicolumn{1}{c|}{192}            & \bf{0.216}         & \bf{0.257}        & 0.307          & 0.367         & 0.597      & 0.598        & 0.245          & 0.285         & 0.276      & 0.336        & 0.673          & 0.597    & {0.219} &0.278        \\
\multicolumn{1}{c|}{}                                   & \multicolumn{1}{c|}{336}            & \bf{0.278}         & \bf{0.293}        & 0.359          & 0.395         & 0.578      & 0.523        & 0.321          & 0.338         & 0.339      & 0.380        & 0.634          & 0.592 &{0.280} &0.322       \\
\multicolumn{1}{c|}{}                                   & \multicolumn{1}{c|}{720}           & \bf{0.341}         & \bf{0.344}        & 0.419          & 0.428         & 1.059      & 0.741        & 0.414          & 0.410         & 0.403      & 0.428        & 0.942          & 0.723    & {0.360} & 0.388      \\ \midrule
\multicolumn{1}{c|}{\multirow{4}{*}{ILI}}                                   & \multicolumn{1}{c|}{24}            & \bf{2.113}         & \bf{0.899}        & 3.483          & 1.287         & 5.764         & 1.677        & 2.294         & 0.945        & 3.228         & 1.660        & 5.800        & 1.693  &3.041 &1.186     \\
\multicolumn{1}{c|}{}           & \multicolumn{1}{c|}{36}            & {1.833}         & \bf{0.800}        & 3.103          & 1.148         & 4.755      & 1.467        & \bf{1.825}          & 0.848         & 2.679      & 1.080        & 6.043          & 1.733   &3.406 &1.232         \\
\multicolumn{1}{c|}{}                                   & \multicolumn{1}{c|}{48}            & \bf{2.012}         & \bf{0.862}        & 2.669          & 1.085         & 4.763      & 1.469        & 2.010          & 0.900         & 2.622      & 1.078        &6.213          & 1.763    &3.459 &1.221     \\
\multicolumn{1}{c|}{}                                   & \multicolumn{1}{c|}{60}           &  \bf{2.114}        & \bf{0.896}        & 2.770          & 1.125        & 5.264      & 1.564        & {2.178}          & 0.963         & 2.857      & 1.157        & 6.531          & 1.814    &3.640 &1.305      \\ \midrule
\multicolumn{1}{c|}{\multirow{4}{*}{Exchange}}               & \multicolumn{1}{c|}{96}            & \bf{0.086}         & \bf{0.205}        & 0.197          & 0.323         & 0.847         & 0.752        & 0.111         & 0.237        & 0.148         & 0.271        & 0.852        & 0.780   &0.246 &0.392    \\
\multicolumn{1}{c|}{}                                   & \multicolumn{1}{c|}{192}           & \bf{0.190}         & 0.304        &   0.300         &  0.369         & 1.204    & 0.895        &  0.219          & 0.335          & 0.271    & \bf{0.280}        &  0.993          & 0.858   &0.889 &0.720         \\
\multicolumn{1}{c|}{}                                   & \multicolumn{1}{c|}{336}           & \bf{0.370}              & \bf{0.431}             & 0.509          & 0.524         & 1.672      & 1.036        &   0.421            &   0.476           & 0.460        & 0.500        &   1.240           &    0.958   &1.375 &0.935      \\
\multicolumn{1}{c|}{}                                   & \multicolumn{1}{c|}{720}           & \bf{0.901}              & \bf{0.710}             & 1.447          & 0.941            & 2.478         & 1.310        & 1.092         & 0.769        & 1.195         & 0.841        & 1.711        & 1.093  &1.978 &1.175     \\ \bottomrule

\end{tabular}
\caption{ The input sequence length is set to 36 for ILI and 96 for the others. Multivariate results with predicted length as \{96, 192, 336, 720\} on the six datasets and \{24, 36, 48, 60\} on the ILI dataset, lower scores are better. Metrics are averaged over 5 runs, best results are highlighted in bold.}
\label{tab:forecasting_result}
	\end{center}
\end{table*}
\section{Model Training}
As mentioned in \cite{cao2020spectral,wen2022transformers}, the prediction-based model is expert in capturing the periodic information of the MTS, while the reconstruction-based model tends to explore the global distribution of the MTS. To combine the complementary advantages of them for facilitating the representation capability of MTS,
we formulate the optimization function as the combination of both prediction and reconstruction losses and define the marginal likelihood as
\begin{equation}\label{margin likelihood} \textstyle
\small
\begin{split}
&P\left(\mathcal{D} \mid\left\{\boldsymbol{W}^{(l)}\right\}_{l=1}^{L}\right)=\int\left[\prod_{t=1}^{T} p\left(\boldsymbol{x}_{t, n} \mid \boldsymbol{z}_{t, n}^{(1)}\right) \right.\\
& +\left[\prod_{l=1}^{L} \prod_{t=1}^{T} p\left(\boldsymbol{z}_{t, n}^{(l)} \mid \boldsymbol{z}_{t, n}^{(l+1)}\right)\right] \\
& +\left.\left[\prod_{l=1}^{L} p \left(\boldsymbol{x}_{T, n} \mid \boldsymbol{h}_{1: T-1, n}^{(l)} \right)\right]\right] d \boldsymbol{z}_{1: T, n}^{1: L} \\
\end{split}
\end{equation}
where the first and the second terms are reconstruction and prediction loss separately.
Similar to VAEs, with the inference network and variational distribution in Eq.~\eqref{inference}, the optimization objective of HTV-Trans can be achieved by maximizing the evidence lower bound (ELBO) of the log
marginal likelihood, which can be computed as
\setcounter{MaxMatrixCols}{20}
\begin{equation} \label{ELBO} \textstyle
\small
\begin{split}
&\mathcal{L}=\sum_{n=1}^{N}\left[\sum_{t=1}^{T} \mathbb{E}_{q\left(\boldsymbol{z}_{t, n}^{(1)}\right)}\left[\ln p\left(\boldsymbol{x}_{t, n} \mid \boldsymbol{z}_{t, n}^{(1)}\right)\right]\right. \\
& +\gamma \mathbb{E}_{q\left(\boldsymbol{z}_{t, n}^{(1)}\right)}\left[\ln p\left(\boldsymbol{x}_{T, n} \mid h_{1: T-1, n}^{(1)}\right)\right] \\
& \left.-\sum_{t=1}^{T} \sum_{l=1}^{L} \mathbb{E}_{q\left(\boldsymbol{z}_{t, n}^{l}\right)}\left[\ln \frac{q\left(\boldsymbol{z}_{t, n}^{(l)} \mid \boldsymbol{x}_{t, n}\right)}{p\left(\boldsymbol{z}_{t, n}^{(l)} \mid \boldsymbol{z}_{t, n}^{(l+1)}\right)}\right]\right] \\
\end{split}
\end{equation}
where $\gamma > 0$ is a hyper-parameter to balance the prediction and the reconstruction losses, which is chosen by grid search on the validation set. The detailed procedures of the optimization of HTV-Trans are summarized in Appendix.

\section{Experiments}
We conduct extensive experiments to prove the effectiveness of our proposed model. Code is available at \url {https://github.com/flare200020/HTV_Trans}.

{\subsection{Datasets and Set Up}}
We evaluate the effectiveness of our model on seven datasets for forecasting, including ETTh1, ETTh2, ETTm1, ETTm2, Illness, Weather and Exhcange-rate \cite{liu2022nonstationary}.  
The results are either quoted from the original papers or reproduced with the code provided by the authors. The way of data preprocessing is the same as \cite{liu2022nonstationary}. 
The summary statistics of these datasets and other implementation details are described in Appendix.




\subsection{Forecasting Main Results}
We deploy two widely used metrics, Mean Absolute Error (MAE) and Mean Square Error (MSE) \cite{zhou2021informer} to measure the performance of forecasting models. Six popular state of the art methods are compared here, including Crossformer \cite{zhang2023crossformer}, Non-stationary transformer \cite{liu2022nonstationary}, Fedformer \cite{zhou2022fedformer}; Pyraformer \cite{liu2021pyraformer};
 Informer \cite{zhou2021informer} and Autoformer \cite{wu2021autoformer}. We note that the experiment settings used here are the same as Non-stationary transformer mentioned \cite{liu2022nonstationary}.
 Table \ref{tab:forecasting_result} presents the overall prediction performance, which is the average MAE and MSE on five independent runs, and the best results are highlighted in boldface. Evaluation results demonstrate that our proposed method outperforms other Transformer-based approaches in most settings, particularly in long-term forecasting tasks. This suggests the effectiveness of our model structure in modeling complex and long-range temporal dependencies. Our method achieves superior performance on almost all datasets, suggesting its ability to capture non-deterministic and non-stationary complex temporal dependencies in MTS, leading to expressive representations and promising prediction outcomes.

\subsection{Ablation Study}
\begin{table*}[!t]
        \center
\begin{tabular}{cl|cl| c c c c c c c c c}
\toprule
\multicolumn{2}{c|}{\multirow{3}{*}{Architecture}}                                                                 & \multicolumn{2}{c|}{\multirow{3}{*}{\begin{tabular}[c]{@{}c@{}}Optimization\\ Objective\end{tabular}}} & \multicolumn{9}{c}{Multivariate time series forecasting}                                                                                               \\ \cmidrule{5-13} 
\multicolumn{2}{c|}{}                                                                                              & \multicolumn{2}{c|}{}                                                                                  & \multicolumn{3}{c}{ETTH2}                        & \multicolumn{3}{c}{Weather}                      & \multicolumn{3}{c}{Exchange}                     \\ \cmidrule{5-13} 
\multicolumn{2}{c|}{}                                                                                              & \multicolumn{2}{c|}{}                                                                                  & 96             & 192            & 336            & 96             & 192            & 336            & 96             & 192            & 336            \\ \midrule
\multicolumn{2}{c|}{Transformer}                                                                                   & \multicolumn{2}{c|}{Prediction}                                                                        & 0.368          & 0.418          & 0.439          & 0.230          & 0.286          & 0.326          & 0.246          & 0.320          & 0.461          \\ \midrule
\multicolumn{2}{c|}{\multirow{2}{*}{\begin{tabular}[c]{@{}c@{}}Transformer with \\ traditional HVAE\end{tabular}}} & \multicolumn{2}{c|}{Prediction}                                                                        & 0.367          & 0.417          & 0.433          & 0.232          & 0.275          & 0.325          & 0.249          & 0.324          & 0.479          \\ \cmidrule{3-13} 
\multicolumn{2}{c|}{}                                                                                              & \multicolumn{2}{c|}{Combine}                                                                           & 0.362          & 0.416          & 0.433          & 0.228          & 0.270          & 0.325          & 0.241          & 0.319          & 0.473          \\ \midrule
\multicolumn{2}{c|}{\multirow{2}{*}{\begin{tabular}[c]{@{}c@{}}Transformer with\\ our HTPGM\end{tabular}}}        & \multicolumn{2}{c|}{Prediction}                                                                        & 0.367          & 0.414          & 0.437          & 0.230          & 0.272          & 0.334          & 0.249          & 0.322          & 0.486          \\ \cmidrule{3-13} 
\multicolumn{2}{c|}{}                                                                                              & \multicolumn{2}{c|}{Combine}                                                                           & \textbf{0.338} & \textbf{0.391} & \textbf{0.405} & \textbf{0.223} & \textbf{0.257} & \textbf{0.293} & \textbf{0.205} & \textbf{0.304} & \textbf{0.431} \\ \bottomrule
\end{tabular}
\caption{Ablation study of HTV-Trans on forecasting tasks. (Transformer denotes the transformer with MLP decoder)}
\label{tab:ablation-study}
\end{table*}
\begin{figure*}[!t]
\centerline{\includegraphics[width=0.95\linewidth,height=0.18\linewidth]{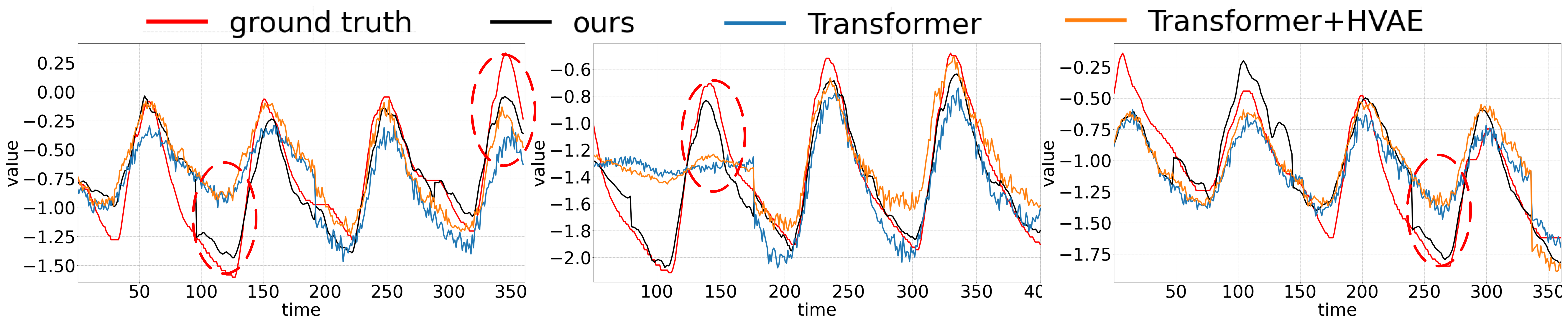}}
\caption{Visualizations on ETTm2 dataset given by different models}
\label{fig:predict-visualization}
\end{figure*}
\begin{figure*}[!t]
\centerline{\includegraphics[width=0.95\linewidth,height=0.2\linewidth]{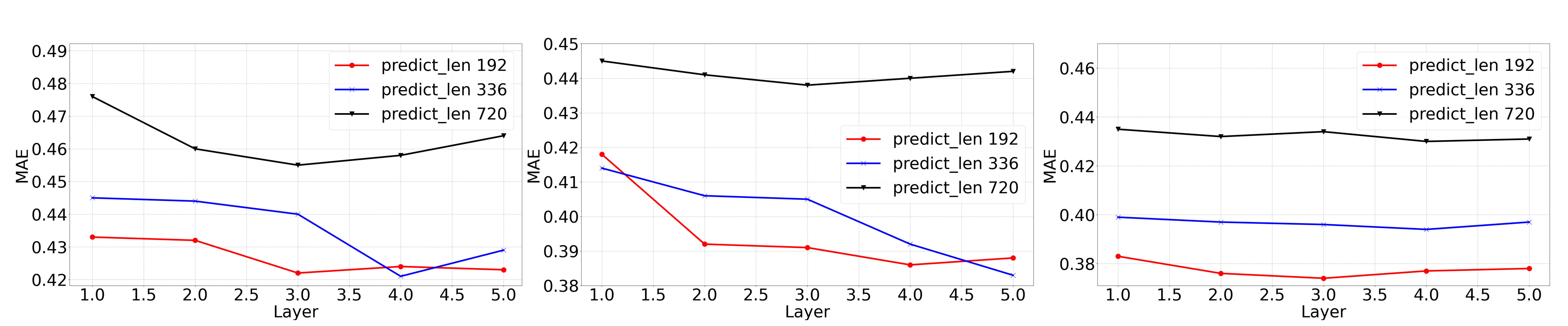}} 
\caption{The effectiveness evaluation of hierarchical architecture on ETTh1(left), ETTh2(middle) and ETTm1(right) dataset.} 
\label{fig:hierarchical_result}
\end{figure*}

{\bf{Prediction quality evaluation:}}
To investigate the contributions of each module in our proposed framework, wo conduct the ablation study by comparing the prediction results of Transformer, Transformer with traditional {\bf H}ierarchical VAE (HVAE), and our HTV-Transformer on the ETTm2 dataset and visualize the results in Figure \ref{fig:predict-visualization}. We find that HTPGM enhances the non-stationary and non-deterministic forecasting ability of Transformers from different angles significantly. Obviously, as demonstrated in Figure \ref{fig:predict-visualization}, traditional Transformer tends to produce output series with high stationarity and volatility, neglecting the non-deterministic and non-stationary nature of real-world data. When the transformer combines the traditional HVAE, the prediction tends to be smooth, but it still fails to capture the non-stationarity of real-world time series due to its poor ability to reconstruct the original series. Unlike the traditional HVAE, the dynamic variable $h$ is introduced in the generative process in our model to help the hidden layer $i$ better learn the distribution of input time series. Besides that, the incorporation of dynamic prior information focuses on aligning the statistical properties of each input series, which helps the Transformer generalize to the whole distribution of data. Both of them play a huge role in our HTPGM. With the incorporation of HTPGM, the model can consider the non-stationary change of real-world time series, leading to improved prediction accuracy for detailed series variation as shown in red circle, which is crucial in real-world time series forecasting tasks.\\\
{\bf{HTPGM architecture evaluation:}}
To further understand the role of HTPGM in our model, we perform an ablation study examining the impact of the hierarchical time series variational scheme, and the combined optimization objective. The results on forecasting tasks are presented in Table~\ref{tab:ablation-study} and Fig.~\ref{fig:hierarchical_result}. We choose Mean Absolute Error (MAE) to measure the performance of forecasting models in Table~\ref{tab:ablation-study} and Fig.~\ref{fig:hierarchical_result}.
As shown in the tables, it is obvious that all structural components contribute to the performance of the framework. In particular, the incorporation of a powerful variational generative module for Transformer leads to a significant improvement in performance, which is proved in Table~\ref{tab:ablation-study}, enabling the model to extract robust and non-stationary representations of MTS with complex distribution. The output of transformer $h$ and dynamic prior information modeling into the generative process further improves the generative capacity of the model and leads to better performance on the forecasting tasks. Furthermore, as shown in the second column of Table~\ref{tab:ablation-study}, the ablation study of the combined optimization objective means the latent variable $z$ can learn the effective and expressive representations of time series. Comparison of the five layers results in Fig~\ref{fig:hierarchical_result} demonstrates the effectiveness of the hierarchical structure. It is because the different layers include different scale robust and non-stationary information which help the representation to be multi-scale. These results verify the efficacy and necessity of each module in our design.
\begin{figure}[!t]
\centerline{\includegraphics[width=1\linewidth,height=0.5\linewidth]{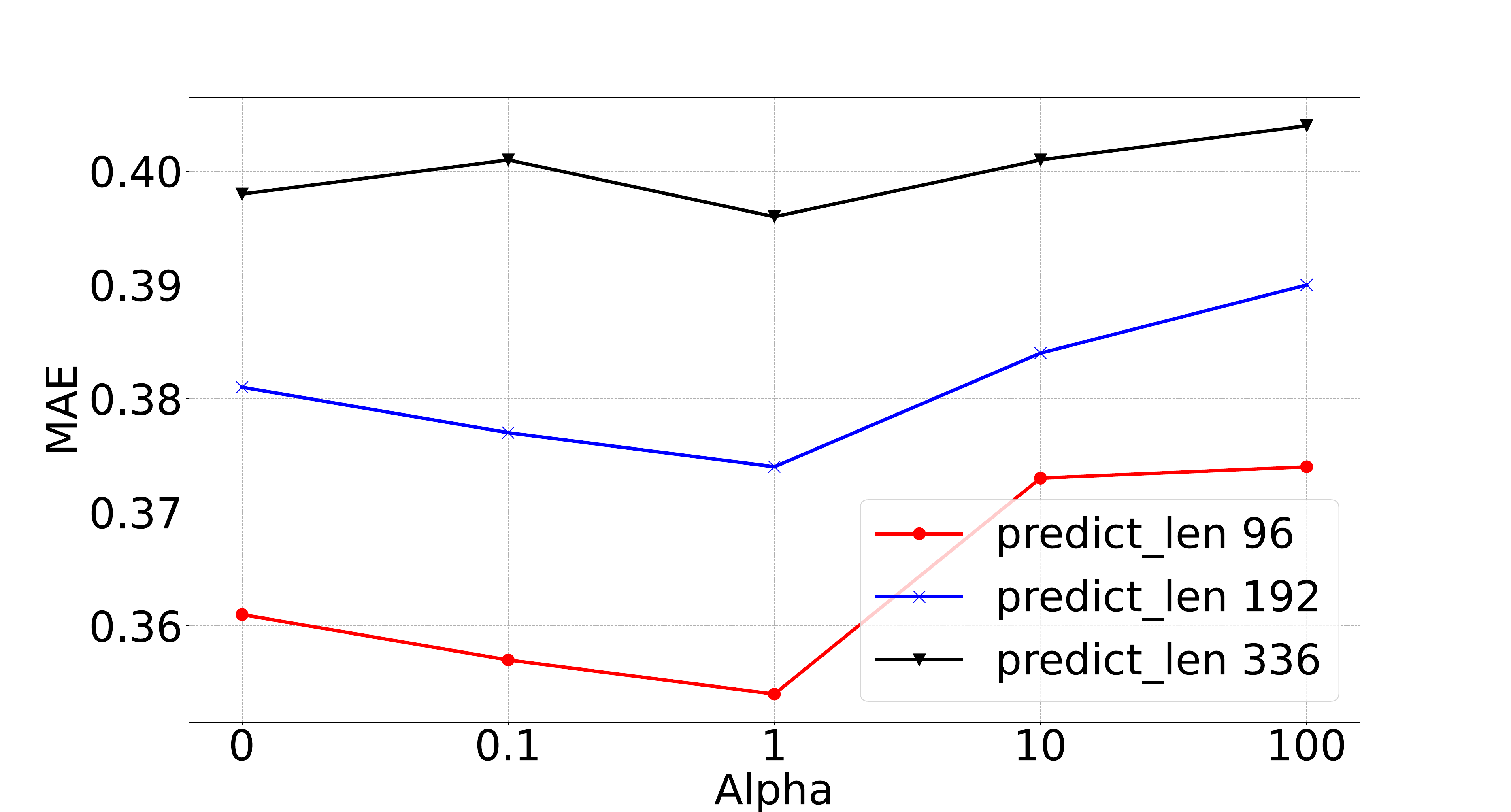}} 
\caption{The effectiveness of parameter $\alpha$ on dataset ETTm1 with different predict length.} 
\label{fig:balance_study}
\end{figure}

\subsection{Balance between Stationary and Non-stationary Information}
As discussed in Sec. 3, we combine normalized $\xv$ and latent representation $\zv$ of original input on HTV-Trans, which enables our models to recover the non-stationary information efficiently and introduce a parameter $\alpha$ to balance the effect of them. Here, we evaluate the influence of
 $\alpha$ to MTS forecasting, the results are reported in Fig.~\ref{fig:balance_study}. As we can see, excessively small and large $\alpha$ will lead to weaker performance,
illustrating the effectiveness of the method that we proposed to recover the non-stationary information. It is because when $\alpha$ is smaller the the non-stationary information is lower, and $\alpha$ is bigger means the the non-stationary information is higher. Only the appropriate value of $\alpha$ can bring suitable non-stationary information. 

\section{Conclusion}

In this paper, we propose a novel approach to solve the over-stationarization problem for multivariate time series forecasting tasks, named HTV-Trans, which consists of a HTPGM module and a transformer that is able to capture non-stationary, non-deterministic, and long-distance temporal dependencies.
To achieve efficient optimization, we introduce an autoencoding variational inference scheme with a combined prediction and reconstruction loss. The HTV-Trans model is able to extract robust and intrinsic non-stationary representations of multivariate time series, which allows it to outperform other models in forecasting tasks. Empirical results on MTS forecasting tasks demonstrate the effectiveness of the proposed model. 
\newpage
\section*{Acknowledgements}
This work was supported in part by the stabilization support of National Radar Signal Processing Laboratory under Grant (JKW202X0X) and National Natural Science Foundation of China (NSFC) (6220010437). The work of Bo Chen acknowledges the support of the National Natural Science Foundation of China under Grant U21B2006; in part by Shaanxi Youth Innovation Team Project; in part by the Fundamental Research Funds for the Central Universities QTZX23037 and QTZX22160; in part by the 111 Project under Grant B18039.

\bibliography{aaai24}
\end{document}